\documentclass[journal]{IEEEtran}
\usepackage{color, xcolor}
\usepackage{url}
\usepackage{times}
\usepackage{helvet}
\usepackage{courier}
\usepackage{verbatim}
\usepackage{hyperref}
\usepackage{graphicx}
\usepackage{comment}
\usepackage{amsmath,amssymb} 
\usepackage{pifont}

\usepackage{CJK}
\usepackage[top=2cm, bottom=2cm, left=2cm, right=2cm]{geometry}
\usepackage{algorithm}
\usepackage{algorithmicx}
\usepackage{algpseudocode}
\usepackage{amsmath}
\usepackage{indentfirst}
\usepackage{booktabs}
\usepackage{multirow}

\floatname{algorithm}{算法}

\newcommand{\yz}[1]{{\color{black} #1}}

\newcommand{\xf}[1]{{\color{black} #1}}

\newcommand{\xingl}[1]{{\color{black} #1}}

\ifCLASSINFOpdf
   \usepackage[]{graphicx}
\else 
\fi

\ifCLASSOPTIONcompsoc
 \usepackage[caption=false,font=normalsize,labelfont=sf,textfont=sf]{subfig}
\else
 \usepackage[caption=false,font=footnotesize]{subfig}
\fi

\begin{document}
%
%
\title{Video Text Tracking with A Spatio-temporal Complementary \xf{Model}}

\author{ Yuzhe Gao*,
         Xing Li*,
         Jiajian Zhang*,
         Yu Zhou,
         Dian Jin,
         Jing Wang,
         Shenggao Zhu,
         Xiang Bai
         
        
\thanks{*Equal contribution.}
\thanks{Corresponding author : Yu Zhou.}
\thanks{This work was supported by the Major Project for New Generation of AI under Grant No. 2018AAA0100400, the National Natural Science Foundation of China 61703049 and the Natural Science Foundation of Hubei Province of China under Grant 2019CFA022. }
\thanks{Y. Gao, X. Li, J. Zhang, Y. Zhou, D. Jin, X. Bai are with School of Electronic Information and Communications, Huazhong University of Science and Technology, Wuhan 430074, China (e-mail: \{yuzhegao, xingl, jiajianzhang, yuzhou, jindian, xbai\}@hust.edu.cn.)}
\thanks{J. Wang, S. Zhu are with Huawei Technologies Co., Ltd., China, (e-mail: \{wangjing105, zhushenggao\}@huawei.com).}}

\maketitle

\begin{abstract}

Text tracking is to track multiple texts in a video,
and construct a trajectory for each text.
Existing methods tackle this task by utilizing the tracking-by-detection framework,
i.e., detecting the text instances in each frame and associating the corresponding text instances in consecutive frames.
We argue that the tracking accuracy of this paradigm is severely limited in more complex scenarios, e.g., owing to motion blur, etc., the missed detection of text instances causes the break of the text trajectory.
In addition, different text instances with similar appearance are easily confused,
leading to the incorrect association of the text instances.
To this end,
a novel spatio-temporal complementary text tracking \xf{model} is proposed in this paper.
We leverage a Siamese Complementary Module to fully exploit the continuity characteristic of the text instances in the temporal dimension,
which effectively alleviates the missed detection of the text instances, 
and hence ensures the completeness of each text trajectory.
We further integrate the semantic cues and the visual cues of the text instance into a unified representation
via a text similarity learning network, 
which supplies a high discriminative power in the presence of text instances with 
similar 
appearance,
and thus avoids the mis-association between them.
Our method achieves state-of-the-art performance on several public benchmarks.
The source code is available at \href{https://github.com/lsabrinax/VideoTextSCM}{\textcolor{magenta}{https://github.com/lsabrinax/VideoTextSCM}}.

\end{abstract}

\begin{IEEEkeywords}
Video Text Tracking, Video Text Detection, Online Tracking.
\end{IEEEkeywords}

\IEEEpeerreviewmaketitle

\section{Introduction}
\IEEEPARstart{N}{owadays},
the visual text exists widely in the scenarios of our daily life.
Since most videos contain text,
detecting and tracking visual text from a video is a significant step in many application,
such as video content review \cite{karatzas2015icdar,liang2015multi},
road signs understanding \cite{reddy2020roadtext(roadtext-1k),minetto2011snoopertrack(minetto)},
video retrieval \cite{fragoso2011translatar(temp-matching_1),sun2017semantic},
automatic drive \cite{YuZhou-IJPRAI2014,FengXue-ICRA2019-TOD,FengXue-TIP2020-TOD,Jianxiang-ICCV2017-OLP,RuiLu-ICCV2019-OFNet,FengXue-PR2021-BSNet,ZheLiu-AAAI2019-TANet},
etc.

Most of the text tracking methods follow the paradigm of tracking-by-detection \cite{yin2016text,YuZhou-IJCV2016-SFVT}, which includes two steps:
detecting the texts in each frame and associating the detected text instances in consecutive frames.
As a result,
the locations of the same text instance in multiple frames form a text trajectory.
Several traditional methods \cite{ minetto2011snoopertrack(minetto), li2000automatic, na2010effective(sift), tanaka2007autonomous(DCT),wang2018scene(backgroundcues)} utilize the conventional detector to locate the text instances and design hand-crafted features to represent the text instances.
Recently,
benefiting from the powerful representation of deep-learning technology,
the text trackers \cite{cheng2020free, yang2017tracking} earn more powerful detectors and stronger discriminative ability to text instances,
thus significantly boosting the tracking accuracy.
Yu \emph{et al.} \cite{yu2019end(baidu)} try to learn the feature embedding in an online association manner.
Cheng \emph{et al.} \cite{cheng2019you(YORO)} and  \cite{cheng2021free(FREE)} propose a tracker for the text instances by using the metric-learning method.

\begin{figure}[t]
    \centering
    \includegraphics[width=1.0\linewidth]{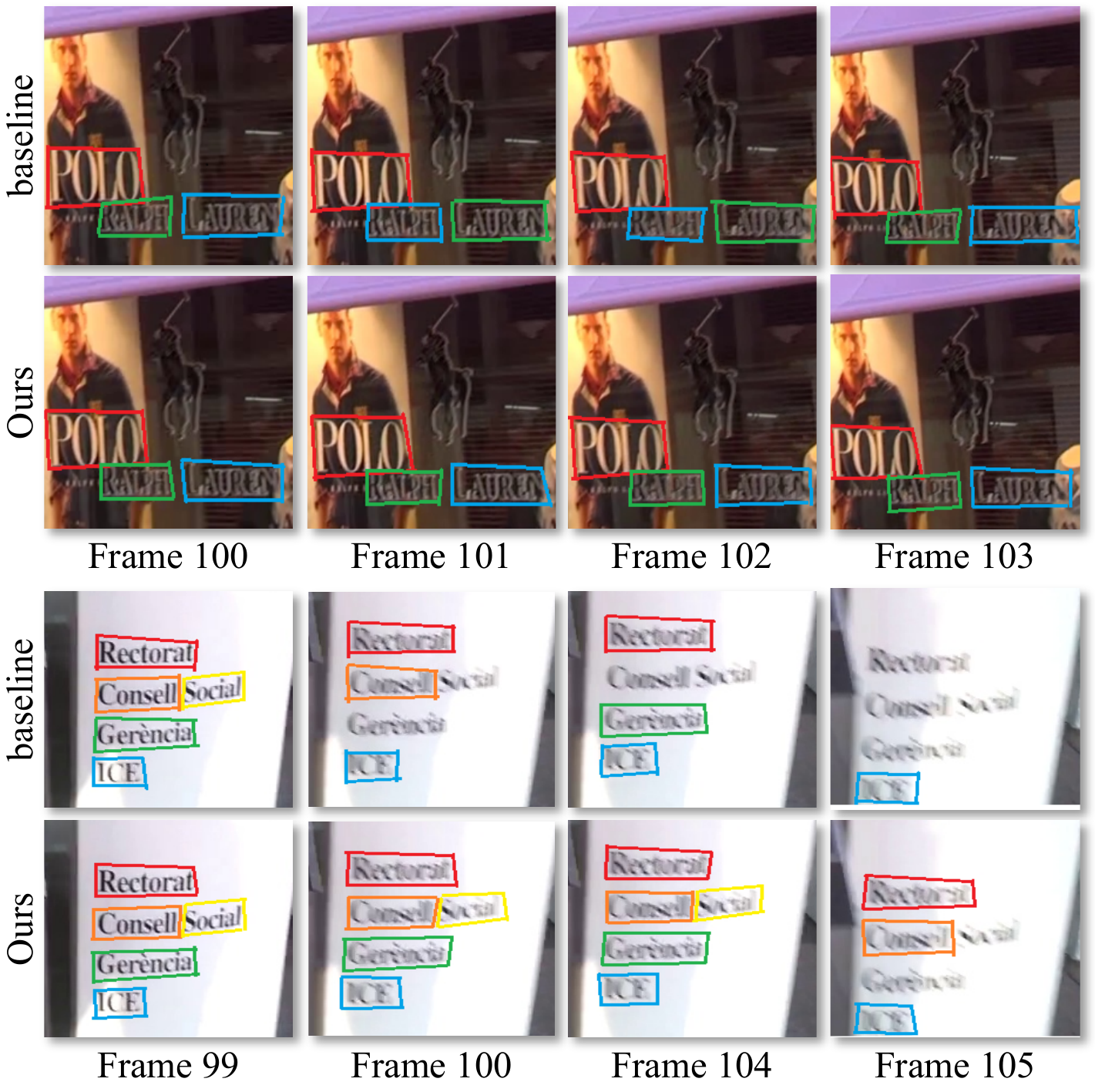}
    \caption{
    The comparative results of text tracking result on ICDAR 2015 Video Dataset. 
    Each row shows the detection and tracking results for four frames in a video, where the text instances with a same color belong to the same trajectory.
    The $1$-st and $3$-rd row are from the baseline methods.
    The $2$-nd and $4$-th are from ours.
    }
    \label{fig:problem vis}
\end{figure}

For these tracking-by-detection approaches,
the association step fully depends on the first step, i.e.,
detecting text in the spatial dimension.
Therefore,
missed detection caused by several physical phenomena,
such as motion blur and various illumination,
greatly degrades the tracking accuracy.
As shown in the third row of Fig.\ref{fig:problem vis},
the text instances marked in yellow, orange, and green are missed by the detector due to motion blur.
Besides,
different texts with similar 
appearances are frequently confused in text association,
leading to the incorrect association of the text instances.
As shown in the first row of Fig.\ref{fig:problem vis},
the similar appearance and style of text cause the ID switch between two text instances marked in blue and green.



To address these issues,
we propose a spatio-temporal complementary model in this paper.
First,
we leverage a Siamese Complementary Module (SCM) to exploit the continuity characteristic of the text instances in the temporal dimension
so as to suppress the missed detection of the text instances.
Therefore, the proposed framework increases the completeness of each text trajectory.
Second,
we further propose a text similarity learning network to integrate the semantic cues and the visual cues of the text instance into a unified representation.
By extracting the semantic feature
encoding the character order of text instances,
we obtain a discriminative representation to the text instances with similar appearance,
thus avoiding the mis-association between text instances.
As shown in the second and fourth rows of Fig.\ref{fig:problem vis},
our method correctly associates the text instances and detects the missing text instances effectively.
Finally, we achieve state-of-the-art performance on the text tracking benchmark, which proves the superiority and effectiveness of our approach.



The contributions of our method are summarized as follows:

\begin{itemize}
\item We propose a spatio-temporal complementary strategy to improve the performance of text localization. 
A Siamese Complementary Module is leveraged to mining the continuity of the temporal dimension, thus localizing the missing text instance.
\item A text similarity learning network is proposed to extract both the visual and semantic features to construct a unified representation, which has a discriminative representation ability for the text instances with a similar appearance.
\end{itemize}

\section{Related Work}

In this section, we review the \textit{Text Detection in Static Images} and \textit{Text Tracking in Video}.
Besides, the \textit{Siamese Network Architecture}, which is related to our method, is also introduced.

\label{Related work}
\subsection{Text Detection in Static Images}
\label{text detection in static images}
Traditional text detection methods mainly based on Connect Components Analysis \cite{epshtein2010detecting(CCA-based_1),huang2013text(CCA-based_2), neumann2010method(CCA-based_3)} 
and Sliding Window \cite{coates2011text(SW-based_1),lee2011adaboost(SW-based_2), wang2011end(SW-based_3)}.
Those methods adopt a hand-crafted pipeline to extract text regions but achieve limited performance.
Recently, with the development of deep learning, great progress has been made in this domain.
CNN-based text detection methods in the static image can be roughly categorized as regression-based and segmentation-based:

\textbf{Regression based} methods such as TextBoxes \cite{liao2017textboxes} modify the anchors and the convolution kernel scale of SSD \cite{liu2016ssd(SSD)} to cope with the extreme aspect ratios. 
RRPN \cite{ma2018arbitrary(RRPN)} and TextBoxes++ \cite{liao2018textboxes++} respectively add rotated anchors and apply quadrilaterals regression to detect multi-oriented text. 
SegLink \cite{shi2017detecting(Seglink)} applies a bottom-up mechanism to predict the text segments and their linkages to handle long oriented text. 
EAST \cite{zhou2017(east)} and DeepReg \cite{he2017(deepReg)} apply pixel-level regression for multi-oriented text instances. 
However, most regression-based methods fail to predict accurate bounding boxes for irregular shapes text.

\textbf{Segmentation-based} methods predict pixel-level text region to handle arbitrary shape text. 
Zhang et al. \cite{long2015fully(FCN)} first attempt to get the text region in an FCN-manner and then group the characters to obtain text instances.
CornerText \cite{lyu2018multi(Corner_text)} detects scene text by localizing corner points of bounding boxes and segments text regions in relative positions.
PSENet \cite{wang2019shape(PSENet)} proposes new post-processing algorithms to separate the near text instances in segmentation-based detectors.
DB \cite{liao2020real(DB)} proposes a Differentiable Binarization module to perform the binarization process in a segmentation network.
Since segmentation can express text of arbitrary shapes well,
we employ DB \cite{liao2020real(DB)} as the text detector to locate texts in each frame.

\subsection{Text Tracking in Video}
Recent text tracking methods mainly design in a tracking-by-detection manner. 
These methods associate the detection results of consecutive frames to generate the final text trajectory. 
The key point of text instance association is to extract the discriminative feature embeddings of different text instances.
Many hand-crafted features like SIFT \cite{lowe2004distinctive(sift)} and HOG \cite{dalal2005histograms(hog)} have been applied to associate the text instances between different frames but gain limited performance.
Hence, metric learning is used in \cite{cheng2019you(YORO)} and  \cite{cheng2021free(FREE)} for video text tracking to obtain more robust visual embedding. 

However, most of these methods only consider the visual information. The semantics information of character sequence, which is significant for distinguishing the different text instances, is rarely discussed by those methods. 
For this reason, we propose a text similarity learning network to extract both the visual and semantic features to produce a robust embedding for text instance association.


\begin{figure*}
    \centering
    \includegraphics[width=1.0\linewidth]{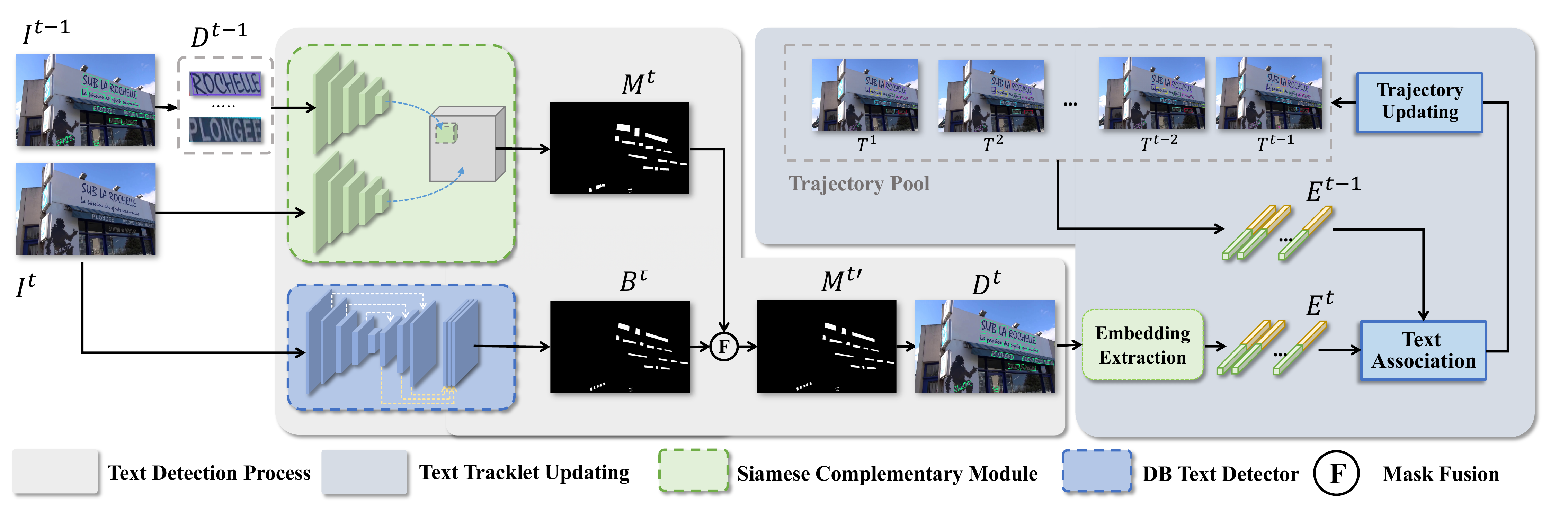}
    \caption{The pipeline of online text tracking. 
    The text trajectories of the previous $t-1$ frame are updated to be associated with the detection text instances in the current frame. 
    }
    \label{fig:online_tracking_pipeline}
\end{figure*}

\subsection{Siamese Network Architecture}

Siamese networks have become a common architecture in various domains, such as single object tracking \cite{bertinetto2016fully(siafc),li2018high(siarpn), wang2019fast(siamask),abdelpakey2019dp_sia_tracking_tip,liang2019local_sia_tracking_tip2}, 
object detection \cite{hu2018deep_sia_det_tip},
representation learning\cite{chen2020siameseRepresent},
face verification \cite{taigman2014deepface}, and one-shot learning \cite{koch2015siamese_oneshot}.
A siamese network consists of twin networks which accept different inputs, then the outputs are joint to extract correlation or comparison.
Specifically, 
the siamese-based trackers like SiameseFC \cite{bertinetto2016fully(siafc)} and SiameseRPN \cite{li2018high(siarpn)} locate the target by the cross-correlation between the deep features of the target template and a search region.
Taigman et al.\cite{taigman2014deepface} apply a siamese network to learn the similarity of the two input person images.
Koch et al. \cite{koch2015siamese_oneshot} perform one-shot classification by learning a convolutional siamese neural networks for verification.
Chen et al. \cite{chen2020siameseRepresent} utilize the siamese networks with a simple detach design for unsupervised visual representation learning.

Our method proposes a Siamese Complementary Module (SCM) to obtain the correlation between the current frame and the missing text of the previous frame, then enhances the missing text of the high-correlation region.

\section{Method}

In this section, we first introduce the overall workflow of our online text tracking approach in Section \ref{online_tracking}. Then, the Siamese Complementary Module (SCM), which re-locates the missed text via the relevance between the consecutive frames, is illustrated in Section \ref{SCModule}. 
Finally, we depict our text similarity learning network in Section \ref{VSENet}, 
which is employed to extract a discriminative embedding from both visual and semantic features. 

\subsection{Online Text Tracking}
\label{online_tracking}

In this paper, 
we formulate the video text tracking problem as text detection and association task in the spatio-temporal domain. 
The pipeline is shown in Fig. \ref{fig:online_tracking_pipeline}.
In particular,
given the text trajectory pool $T^{t-1} = \{\tau_{0}^{t-1},\tau_{1}^{t-1},...,\tau_{M}^{t-1}\}$ tracked from the previous $t-1$ frames.
Each text trajectory $\tau_{i}^{t-1} = \{b_{i}^{m},...,b_{i}^{n}\}_{0 \leq m < n <t}$  belongs to a text instance with the trajectory ID $i$,
and this text first appears in the $m$-th frame and ends in the $n$-th frame. 
In addition, each text instance is parametrized as $b_i^{t} = \left(v_i^{t},e_i^{t},c_i^{t},i \right)$, 
where $v \in \mathbb{R}^{4\times2}$ is the four vertexes of the bounding box, 
$e \in \mathbb{R}^{1 \times 512}$ is the feature representation of the text instance,
$c \in \left[0,1\right]$ is the detection confidence score, 
and $i \in \mathbb{I}$ is the trajectory ID.
In the current frame $I^{t}$,
we leverage the proposed spatio-temporal text localization strategy to locate a set of candidate text instances $D^{t} = \{d_1^{t},d_2^{t},..., d_{N}^{t}\}$,
where $d_i^{t}=(v_i^t, c_i^t)$.

Then, the goal of online text tracking is to assign the trajectory ID in the latest trajectories set $T^{t-1}$ for each detected text instance in $D^{t}$ by text association.
We extract the feature embedding $E^{t} = \{e_0^{t},..., e_{j}^{t}\}$ of each text instance in $D^{t}$.
The embedding distance $\mathcal{D}_{e}$ of the feature embedding is utilized as a similarity metric between two text instances.
Besides, we also employ an IOU distance $\mathcal{D}_{p}$ and a morphological distance $\mathcal{D}_{m}$ for the robust association.
Finally, the text instances in $D^{t}$ with the corresponding trajectory ID are added to the text trajectory pool, resulting in the updated trajectory pool $T^{t}$.

\begin{figure}
    \centering
    \includegraphics[width=1.0 \linewidth]{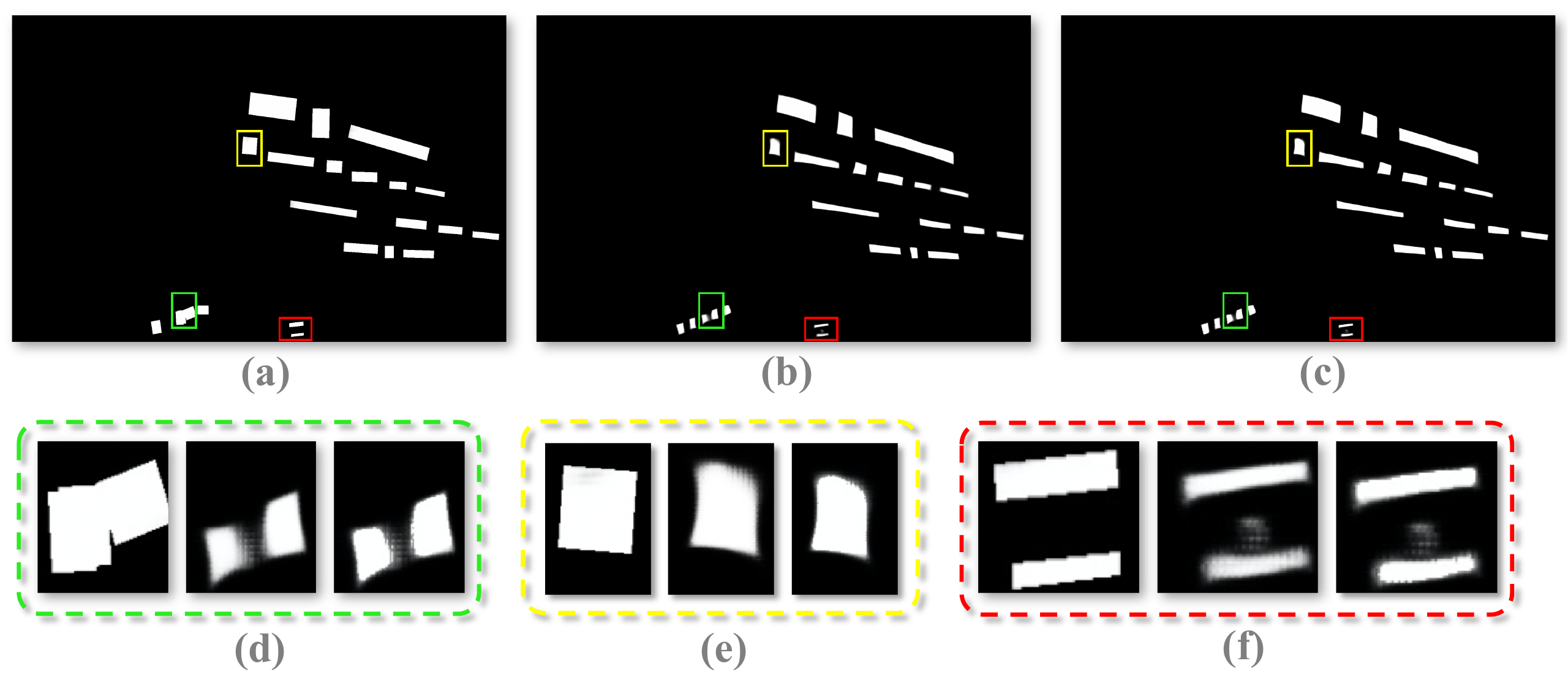}
    \caption{The visualization of the probability map in Spatio-temporal Text Localization. 
    (a) is the binary mask $M^t$ of the SCM.
    (b) is the probability map $B^t$of the text detector.
    (c) is the fused probability map $B^{t'}$.
    Some regions are enlarged for better visualization, as shown in the green, yellow and red boxes.
    In each block of (d) (e) (f), the three image patches come from $M^t$,  $B^t$ and  $B^{t'}$ respectively.  
    }
    \label{fig:prob}
\end{figure}

\textbf{Spatio-temporal Text Localization:}
In order to alleviate the text trajectory break caused by motion blur, illumination change, etc., 
we present a spatio-temporal text localization strategy.
Specifically, for text localization in the spatial domain, 
we leverage DB \cite{liao2020real(DB)} to detect the text in the current frame $I^{t}$.
As depicted in Fig. \ref{fig:online_tracking_pipeline},
$I^{t}$ is fed into the ResNet50 based FPN backbone,
and the last four pyramid features are up-sampled to $\frac{1}{4}$ scale of $I^{t}$.
Then these four features are concatenated, and employed to predict a text probability map {$B^{t}$},
which indicates the probability of valid text areas at each pixel in the input frame $I^{t}$.
In addition, for the text localization in the temporal domain, 
we leverage the text instances $D^{t-1}$ in the last frame $I^{t-1}$
to predict a possible text instance in $I^{t}$ by the Siamese Complementary Module (SCM) presented in Sec. III.B,
resulting a binary mask $M_{t}$ which indicates the location of the text instances of $D^{t-1}$ in current frames .
We use $M^t$ to enhanced the probability map $B^t$ to produce $B^{t'}$, and then binarize $B^{t'}$ by a threshold $h_2$ to obtain $M^{t'}$:
\begin{equation}
\begin{aligned}
 B^{t'} &= M^t[B^{t} >h_1]+B^t \\
 M^{t'} &= B^{t'}>h_2
\end{aligned}
\label{eq: mask fusion}
\end{equation}
where $h_1$, $h_2$ are two threshold.
Firstly, for all pixels in $B^t$ with a probability value greater than $h_1$, we add the value (0 or 1) of this pixel in $M^t$  to it.
This updated probability map is binarized by a threshold $h_2$ to produce the binary masks $M^{t'}$.
Since some text regions have low probability and cause misdetection, 
we boost them by the predicted mask $M^t$.
Hence, if this text instance is located by our SCM and the text region have values of 1 in  $M^t$, the probabilities in $B^t$ are enhanced.
To verify the effectiveness of our proposed SCM, we visualize the probability map $B^t$,$B^{t'}$ and the binary masks $M^t$  in Fig.\ref{fig:prob}.
The weak probabilities in $B^t$ of some text instances are significantly enhanced in $B^{t'}$, as shown in the enlarged green, yellow and red boxes.
We generate the bounding box for each text instance from $M^{t'}$.
Thus, the detection result $D^t$  in the current frame is obtained for text association.

\textbf{Text Association:}
After text localization in $I_t$,
we need to associate each text instance in $D^t$ to a trajectory in $T^{t-1}$.
For each trajectory $\tau_{i}^{t-1} = \{ b_{i}^{m}, ..., b_{i}^{n} \}$ in $T^{t-1}$, we get the text instance $b_{i}^{n}$ which is latest tracked in the previous $t-1$ frames and calculate the distance between all the text instances in $D^t$, resulting in an adjacent distance matrix $A_{dist} \in \mathbb{R}^{M\times N}$ for the subsequent association.
Therefore, a distance is needed to measure the similarity of two texts, and two texts with higher similarity are more likely to belong to the same trajectory.
Our distance measure consists of the following three metrics.

\emph{Text Appearance Distance} $\mathcal{D}_{e}$: 
For each text instance in $D^t$, 
we leverage RoI-Align \cite{he2017mask(mask-r-cnn)} to extract the fixed-size text instance feature,
and then send it to the subsequent text similarity learning network to extract the text feature embedding.
After extracting the embedding $E^t$ of all the text instances in $D^t$,
we obtain the embedding $e_{i}^{n}$ of the latest tracked instance $b_{i}^{n}$ in each trajectory,
and calculate the embedding distance with $E^t$. 
Embedding distance $\mathcal{D}_{e}$  is measured as the $L$2 Euclidean distance between two feature embedding.

\emph{IOU Distance} $\mathcal{D}_{p}$: 
Consider that a text instance is more likely to appear in close locations in adjacent frames, we also measure the distance of two text instances by the similarity of their bounding box, i.e., the Intersection over Union (IOU). 
Hence, the IOU distance $\mathcal{D}_{p}$ is utilized,
as mentioned in \cite{bochinski2017high(IOUtracker)}. 

\begin{figure}[t]
    \centering
    \includegraphics[width=0.7 \linewidth]{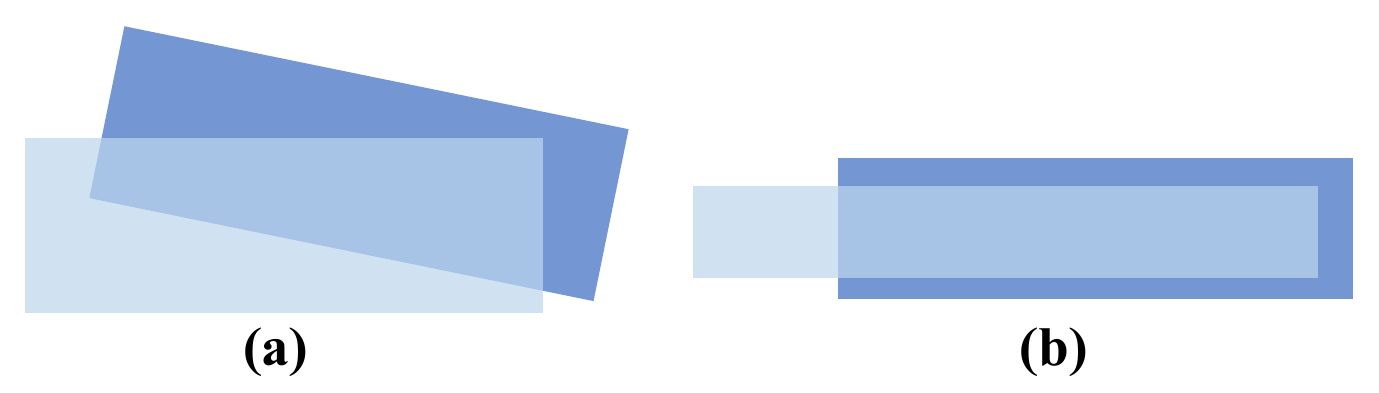}
    \caption{Two examples of high IOU but low morphological similarity. 
    }
    \label{fig:mor_dist}
\end{figure}

\emph{Morphological Distance} $\mathcal{D}_{m}$:
Only IOU distance cannot fully reflect the similarity of the two bounding boxes, e.g., 
the bounding boxes of two text instances that belong to different trajectories have high IOU, 
but their centers, length, width are quite different,
as depicted in Fig.\ref{fig:mor_dist}.
We leverage the morphological distance matrix $\mathcal{D}_{m}$, which measures the similarity of shape and position to alleviate the above phenomenon.
The formulation of $\mathcal{D}_{m}$  is stated as:
\begin{equation}
\begin{split}
\mathcal{D}_{m}\left(r_{i}, r_{j}\right) =& \sigma_{1}( \frac{|x_{i}-x_{j}|+|y_{i}-y_{j}|+|h_{i}-h_{j}|+|w_{i}-w_{j}|}{ \Delta f_{ij}}) \\
&+ \sigma_{2} (|\frac{w_{i}}{h_{i}} - \frac{w_{j}}{h_{j}}|) 
+ \sigma_{3} (|\theta_{i} - \theta{j}|)
\end{split}
\label{eq: morphology dist}
\end{equation}
\noindent where $r_i=(x_i,y_i,w_i,h_i,\theta_i)$  denotes the rectangle bounding box of a text instance $d_i$. 
$(x_i,y_i)$ is the center of the rectangle. $w_i,h_i$ are the width and height, $\theta_i$ is the oriented angle of the bounding box,  
$\Delta f_{ij}$ is the frame interval between the frames of two text instances $d_i$ and $d_j$, 
i.e., when calculating the distance between a text instance in the current frame and a text instance latest tracked in the $n$-th frame, the frame interval $\Delta f_{ij}=|t-n|$.
$\sigma_{1}$, $\sigma_{2}$, $\sigma_{3}$ are the hyper-parameters that control the balance between the three terms in Eq. \eqref{eq: morphology dist}.

The first term in Eq. \eqref{eq: morphology dist} presents the difference of shape and position.
The second term presents the difference in aspect ratio.
The third term presents the difference in box orientation.
Considering that the shape and position of a text instance are more likely to change in different frames, we use the reciprocal of $\Delta f_{ij}$ to balance the first term.

Then we get the overall distance $\mathcal{D}$ as follows:
\begin{equation}
\begin{split}
\mathcal{D} = \alpha\mathcal{D}_{e} + \beta\mathcal{D}_{p} + \gamma\mathcal{D}_{m}  \\ 
\end{split}
\label{eq:overall distance}
\end{equation}
\noindent where $\alpha, \beta, \gamma$  are the weighting factors of each distance.

Thus, we calculate the distance between the text instances in $D^t$ and $T^{t-1}$, resulting in the adjacent distance matrix $A_{dist}$ as follow:
\begin{equation}
    A_{dist} 
    =
    \left[
    \begin{array}{cccc}
        \mathcal{D}_{11} & \mathcal{D}_{12} & \cdots & \mathcal{D}_{1N} \\
        \mathcal{D}_{21} & \mathcal{D}_{22} & \cdots & \mathcal{D}_{2N} \\
        \vdots          & \vdots          & \ddots & \vdots          \\
        \mathcal{D}_{M1} & \mathcal{D}_{M2} & \cdots & \mathcal{D}_{MN} \\
    \end{array}
    \right],
\end{equation}

\noindent where $N$ and $M$ are the number of text instances of $D^{t}$  and $T^{t-1}$ respectively. 
$\mathcal{D}_{ij}$ is the text distance between $d_i$ in $T^{t-1}$ and $d_j$ in $D^{t}$.

Then we perform Hungarian Algorithm on $A_{dist} $ to achieve the minimum distance during text association.
Each text instance $d_{i}$  in the current frame is associated with the $\tau_{j}^{t-1}$  of the minimal distance in $T^{t-1}$.

\textbf{Trajectory Updating:}
If a text instance $d_{i}^t$ is associated with a text trajectory $\tau_{j}^{t-1}$, 
$d_{i}$ is added to the trajectory $\tau_{j}$ to become $b^t_j$.
If a detected text instance $d_{i}^t$ is not associated with any activated text trajectory,
we initialize $d_{i}$  to be a new text trajectory in $T^{t}$ and assign a new trajectory ID.

For a text trajectory $\tau_{j}$ that fails to associate any detected text instance in the current frame, it is marked as lost. 
If a text trajectory is marked as lost in several continuous frames,
we consider that this trajectory is missed and remove it from $T^{t}$.

\begin{figure}[t]
    \centering
    \includegraphics[width=1.0 \linewidth]{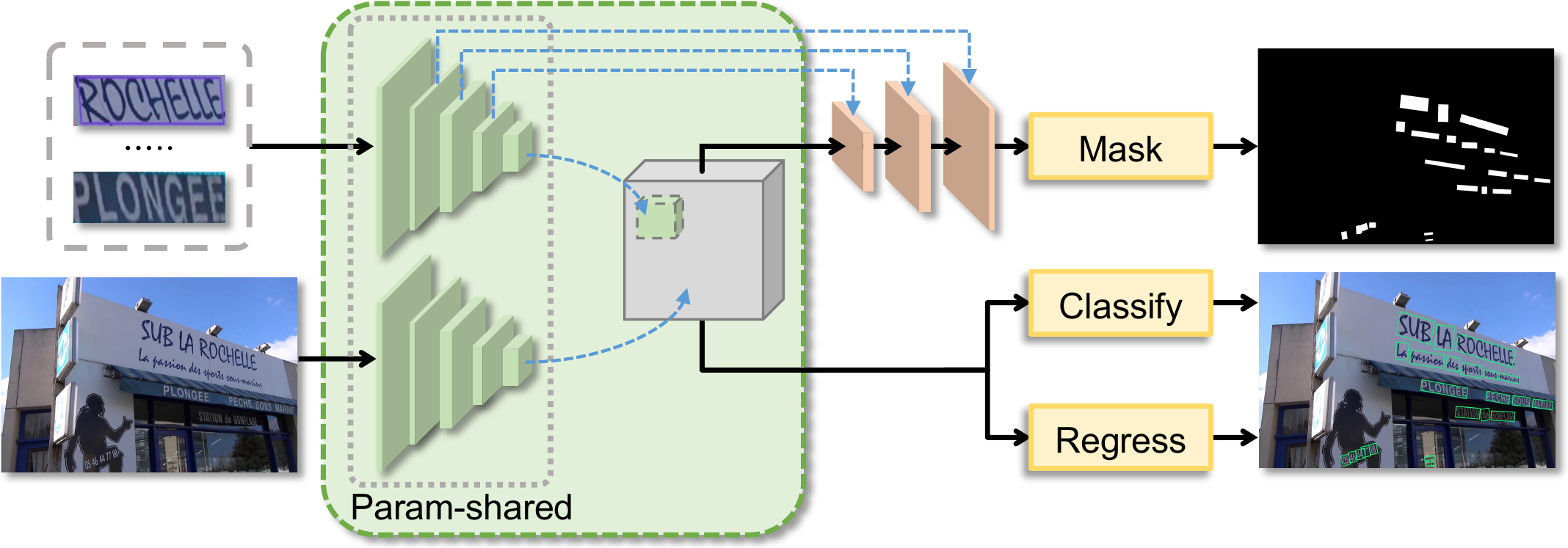}
    \caption{\yz{The architecture of Siamese Complementary Module. }
    }
    \label{fig:SCM}
\end{figure}

\subsection{Siamese Complementary Module}
\label{SCModule}

To detect the missing text instances due to the complex situation as mentioned above,
we design the Siamese Complementary Module (SCM) to locate and track those missed text instances by leveraging the correlation of text instances in consecutive frames.

The architecture of the SCM is illustrated in Fig \ref{fig:SCM}. 
It consists of a parametric-shared siamese network $\phi$ and three branches for text classification, bounding box regression and text mask prediction, denoted as $\phi_c$, $\phi_r$, $\phi_m$ respectively.
For each detected text instances $d_i^{t-1}$ in the last frames, 
we leverage its image patch as a template $\mathcal{Z}_i^{t-1}$, 
and a search region $\mathcal{X}_i^t$ is cropped in the current frame $I^{t}$ with double the size of $\mathcal{Z}_i^{t-1}$.
Both the template  $\mathcal{Z}_i^{t-1}$ and search region  $\mathcal{X}_i^t$ are input to the parametric-shared siamese network $\phi$ and conduct the correlation operation to locate the potentially text instance in the current frame $I^{t}$.
Specifically, the feature $\phi(\mathcal{Z}_i^{t-1})$ of the template is served as a convolution kernel, 
and we perform convolution operation on $\phi(\mathcal{X}_i^{t})$ to obtain a cross-correlation map $h_{\phi}(\mathcal{Z}_i^{t-1}, \mathcal{X}_i^{t})$:

\begin{equation}
\begin{split}
h_{\phi}(\mathcal{Z}_i^{t-1}, \mathcal{X}_i^{t}) = 
\phi(\mathcal{Z}_i^{t-1})  \ \star \  \phi(\mathcal{X}_i^{t})
\end{split}
\label{eq:distance}
\end{equation}
\noindent where $\star $ denotes correlation operation.

We use the subsequent text classification branch and bounding box regression branch to predict the classification score and regress the bounding box.
Further, the mask prediction branch is used to predict a binary mask in each location,
which indicates the detection result located by the text instances.
\yz{Following \cite{wang2019fast(siamask)},
we adopt FPN-like multi-scale features in the mask branch while only use the last correlation map in the text classification branch and the bounding box regression branch.
The reason is that both the bounding box regression branch and the text classification branch are sensitive to the setting of the anchors. 
If we use multi-scale information, we need to set anchors in each scale of the feature, which may greatly \xingl{reduce}
the inference speed \xingl{,while the mask prediction branch does not suffer from this issue.}}
When all the binary mask of text instances in $D^{t-1}$ is obtained, we use logical 'or' operation to fuse them, resulting in a binary mask $M^t$.

\textbf{Model Training:}
All the training images are picked from the video in \emph{ ICDAR 2013} and \emph{ICDAR 2015}\cite{karatzas2015icdar} datasets. 
Following \cite{li2018high(siarpn)},
we crop the patch centering on a text instance from the image to get a template, and the search region patch is cropped with the double size of the template. 
The template and search region patches are resized to fixed $127\times127$ and $255\times255$ respectively to form the training pair.
The loss function is illustrated as follow: 
\begin{equation}
\begin{split}
\mathcal{L} = \lambda_1 L_{cls} + \lambda_2 L_{reg} + \lambda_3 L_{mask}
\end{split}
\label{eq: loss function}
\end{equation}
where $L_{cls}$ is the cross-entropy loss for text classification,
$L_{reg}$ is the smooth $L1$ loss for the bounding box regression, 
$L_{mask}$ is the binary logistic regression loss mentioned in \cite{wang2019fast(siamask)}. 
$\lambda_1,\lambda_2,\lambda_3$ is the hyper-parameters to balance the three parts, set as 1.0, 1.0, 32.0.
The SCM is trained on one machine with a NVIDIA TITAN Xp GPU.
We train the SCM for 8 hours with batch size of 64,
and SGD is adopted to optimize the parameters.
The learning rate is set as a fixed $1 \times 10^{-5}$.

\begin{figure*}
   
   \centering
    \includegraphics[width=1.0 \linewidth]{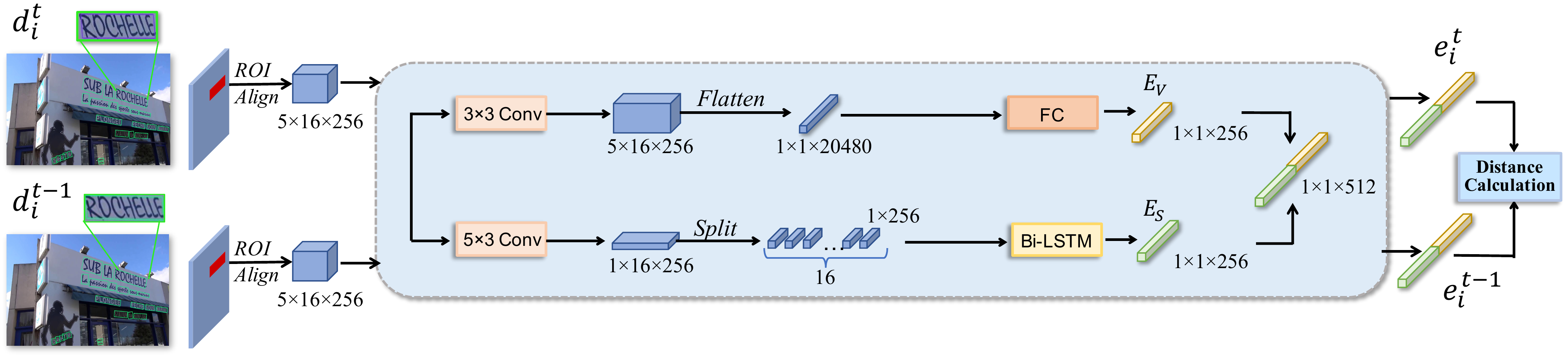}
    \caption{The architecture of the text similarity learning network. 
    It consists of two branches for the visual and semantic embedding, respectively. 
    }
    \label{fig:vs_module}
\end{figure*}

\subsection{Text Similarity Learning Network}
\label{VSENet}

As depicted in Fig. \ref{fig:vs_module},
aiming to measure the distance between two text instances,
we present a text similarity learning network to extract instance embedding for text association.
This network consists of two branches for extracting visual embedding and semantic embedding, respectively.

\emph{Semantic embedding:}
Since the text instances have the specific text meaning represented by the sequential order of characters, conventional convolution cannot derive these sequence-like object's semantic meanings. 
Thus, we convert the text instance feature to a sequence and model the semantic feature by the Long Short-Term Memory (LSTM).
To be specific, 
for each text instance $d_i^t$, 
we obtain its cropped fixed-size  $5 \times 16 \times 256$ feature and adopt a $5 \times 3$ convolution to transfer it into $1 \times 16 \times 256$ feature map.
This feature map is separated into a sequence of length 16 composed of $ 1 \times 256$ feature vectors.
Then the stacked Bi-directional LSTM\cite{shi2016end(CRNN)} is employed to capture the correlation between these feature vectors. 
Finally, the semantic feature is reduced to a sequence of length 16 composed of a $1 \times 16$ feature vector, then concatenated to a 256-dimension vector,
served as the semantic embedding ${e^S}_i^{t}$.

\emph{Visual embedding:}
Aiming to learn a holistic representation of the text instance,
the cropped fixed-size feature of text instance $d_i^t$ is firstly fed into two $3 \times 3$ convolution layers, then flattened to one dimension vector.
A fully connected layer is applied to produce a $1 \times 256$ feature vector, 
served as the visual embedding ${e^V}_i^{t}$.

Both the visual and semantic embedding is a $1 \times 256$ vector, and we concatenate these two embeddings to produce the text instance embedding $e_i^t$ of $ 1 \times 512$.

We utilize the triplet loss \cite{hermans2017defense(reid-tri)} to train the text instance embedding $e_i^t$.
In each iteration, a triplet consists of two instances $a$, $p$ from the same trajectory and an instance $n$ from other trajectories.
Hence, the triplet loss is illustrated as follow:
\begin{equation}
L_{tri} = \sum_{id_{a} = id_{p} \neq id_{n}} \max ( 0, -{(|e_a - e_n| - |e_a-e_p|)} + \mathcal{M})
\label{loss_tri}
\end{equation}
\noindent where $id$ denotes the trajectory identity.
$e_a, e_n, e_p$ denote the instance embedding of $a,p,n$ respectively.
$|\cdot|$ denotes the $L$2 Euclidean distance.
$\mathcal{M}$ is the margin value, set to 1.0.

\emph{Model Training:}
During the training phase, 
two frames are fed into the text similarity learning network.
We sample the image pairs from the videos with random intervals (max to 5 frames) and then resize them to $800\times800$.
All the training data comes from the \emph{ICDAR Video Text 2013} and \emph{ICDAR Video text 2015}\cite{karatzas2015icdar} dataset.
Besides, the hard-mining strategy\cite{hermans2017defense(reid-tri)} is also adopted to select the hard samples for training.
In the training phase, 
This module is also trained on a NVIDIA TITAN Xp for 22 hours with batch size of 1.
RMSprop is used for optimization, and the initial learning rate is set to $10^{-4}$, with a decay rate of 0.3 for every 15 epochs.


\section{Experiments}

In this section, we first introduce several public benchmark datasets and evaluation metrics, and then compare our method with state-of-the-art approaches on the benchmark datasets. Finally, we perform sufficient ablation studies to prove the effectiveness of our design.

\subsection{Experimental settings}
\subsubsection{Datasets}
In this paper, three video text datasets are utilized for evaluation, as illustrated below: 
\vspace{1mm}

\textbf{ICDAR 2013 Video Text Dataset: }
ICDAR 2013 Video Text Dataset\cite{karatzas2015icdar} contains 28 videos lasting from 10 seconds to 1 minute. 
We use 13 long videos for training and 15 short videos for testing. 
These videos are captured by 4 types of cameras in 7 different nature scenes. 

\textbf{ICDAR 2015 Video Text Dataset:}
ICDAR 2015 Video Text Dataset\cite{karatzas2015icdar} is the largest video text dataset.
It contains 49 videos, 
25 videos for training and 24 videos for testing.
The evaluation results of the existing methods are shown on the ranking list of the ICDAR benchmark website.

\textbf{Minetto Dataset:}
Minetto dataset \cite{minetto2011snoopertrack(minetto)} contains 5 videos captured in the outdoor scenario. 
The frame size is $640 \times 480$, and all the videos are used for testing.
\vspace{1mm}

\subsubsection{Evaluation metrics}
For the evaluation of text tracking, We use the same metrics as \cite{ristani2016IDFmetric}. 
These metrics come from the CLEAR-MOT framework \cite{bernardin2008evaluatingCLEAR} for multiple object tracking,
including Multiple Object Tracking Accuracy ($\text{MOTA}_D$),
Multiple Object Tracking Precision ($\text{MOTP}_D$) and Identification F1-score (IDF1).
The suffix 'D' means tracking is applied for detection.
Besides, metrics of Mostly-Matched (MM), Partial-Matched (PM), 
Mostly-Lost (ML) and ID Switch are also adopted.
Specifically, Mostly-Matched, Partial-Matched, and Mostly-Lost represent the degree to which a text instance is correctly tracked.
ID Switch represents that the trajectory ID of a text instance changes during text tracking. 
Besides, we also adopt Precision, recall, and F-measure for the evaluation of text detection.   

\subsubsection{Hyper-parameter setting}
During the online text tracking,
we employ a same setting of hyper-parameters for the experiments in all datasets.
$h_1$, $h_2$ in Eq.\eqref{eq: mask fusion} are set as 0.6, 0.3.
$\sigma_{1}$, $\sigma_{2}$, $\sigma_{3}$ in Eq.\eqref{eq: morphology dist} are set as 0.3, 0.3, 0.7.
$\alpha$, $\beta$, $\gamma$ in Eq.\eqref{eq:overall distance} are set as 0.6, 0.2, 0.2.
More discussions about parameters are given in section \ref{sec:param}.


\begin{table}[htbp]
\renewcommand\arraystretch{1.3}
\centering
\caption{Text tracking results of ICDAR 2015 Video Text Dataset}
\label{tab:comparison on tracking results icdar}
\begin{tabular}{ p{1.6cm} | p{1cm} |  p{1cm} | p{0.8cm} | p{0.5cm} | p{0.5cm} | p{0.5cm} }
\hline
Methods&$\text{MOTA}_D$$\uparrow$&$\text{MOTP}_D$$\uparrow$&IDF1$\uparrow$&MM$\uparrow$&PM$\downarrow$&ML$\downarrow$ \\
\hline
StradVision\cite{rrc.cvc.uab.es} &7.92$\%$&70.17$\%$&25.87$\%$&124&436&1356 \\
Megvii\cite{rrc.cvc.uab.es} &11.02$\%$&66.75$\%$&37.30$\%$&185&612&1119 \\
USTB\cite{rrc.cvc.uab.es} &12.29$\%$&71.78$\%$&21.93$\%$&92&439&1385 \\
AJOU\cite{rrc.cvc.uab.es} &16.44$\%$&72.71$\%$&36.07$\%$&271&458&1187 \\
SRC\-B\ \cite{rrc.cvc.uab.es} &23.09$\%$&68.51$\%$&39.40$\%$&274&481&1161 \\
HIK\_OCR\cite{cheng2021free(FREE)}&43.16$\%$&\textbf{76.78$\%$}&57.92$\%$&702&\textbf{364}&850 \\
\hline
Ours &\textbf{44.07$\%$}&75.19$\%$&\textbf{58.23$\%$}&\textbf{858}&502&\textbf{556} \\
\hline
\end{tabular}
\end{table}

\begin{table}[htbp]
\renewcommand\arraystretch{1.2}
\centering
\caption{Text tracking results of Minetto Dataset}
\label{tab:comparison on tracking results minetto}
\begin{tabular}{c|c|c}
\hline
Method&$\text{MOTA}_D$$\uparrow$&$\text{MOTP}_D$$\uparrow$ \\
\hline
Zuo et al.\cite{7333727} &56.37$\%$&73.07$\%$ \\
Pei et al.\cite{8283802}&57.71$\%$&73.07$\%$\\ 
Yu et al.\cite{yu2019end(baidu)}&81.31$\%$&75.72$\%$ \\
\hline
Ours &\textbf{86.9}&\textbf{83.10} \\
\hline
\end{tabular}
\end{table}

\begin{table}[!htbp]
\renewcommand\arraystretch{1.2}
\centering
\caption{Text detection results comparison on ICDAR 2013 Video Text Dataset}\label{tab:detection result}
\begin{tabular}{c|c|c|c}
\hline
Method&Precision&Recall&F-measure\\
\hline
Khare et al.\cite{khare2015new}& 47.60& 41.40&44.30\\
Zhao et al.\cite{zhao2010text}& 46.30& 47.02&46.65\\
Shivakumara et al.\cite{shivakumara2012multioriented}&51.15&53.71&50.67\\
Yin et al.\cite{yin2013robust}&48.62&54.73&51.56\\
Wang et al.\cite{wang2018scene(backgroundcues)}&58.34&51.74&54.45\\
Yu et al.\cite{yu2019end(baidu)}&80.00&55.20&65.30\\
\hline
Ours &\textbf{82.41}&\textbf{59.22}&\textbf{68.92}\\
\hline
\end{tabular}
\end{table}

\begin{figure*}[htbp]
    \centering
    \includegraphics[width=1.0 \linewidth]{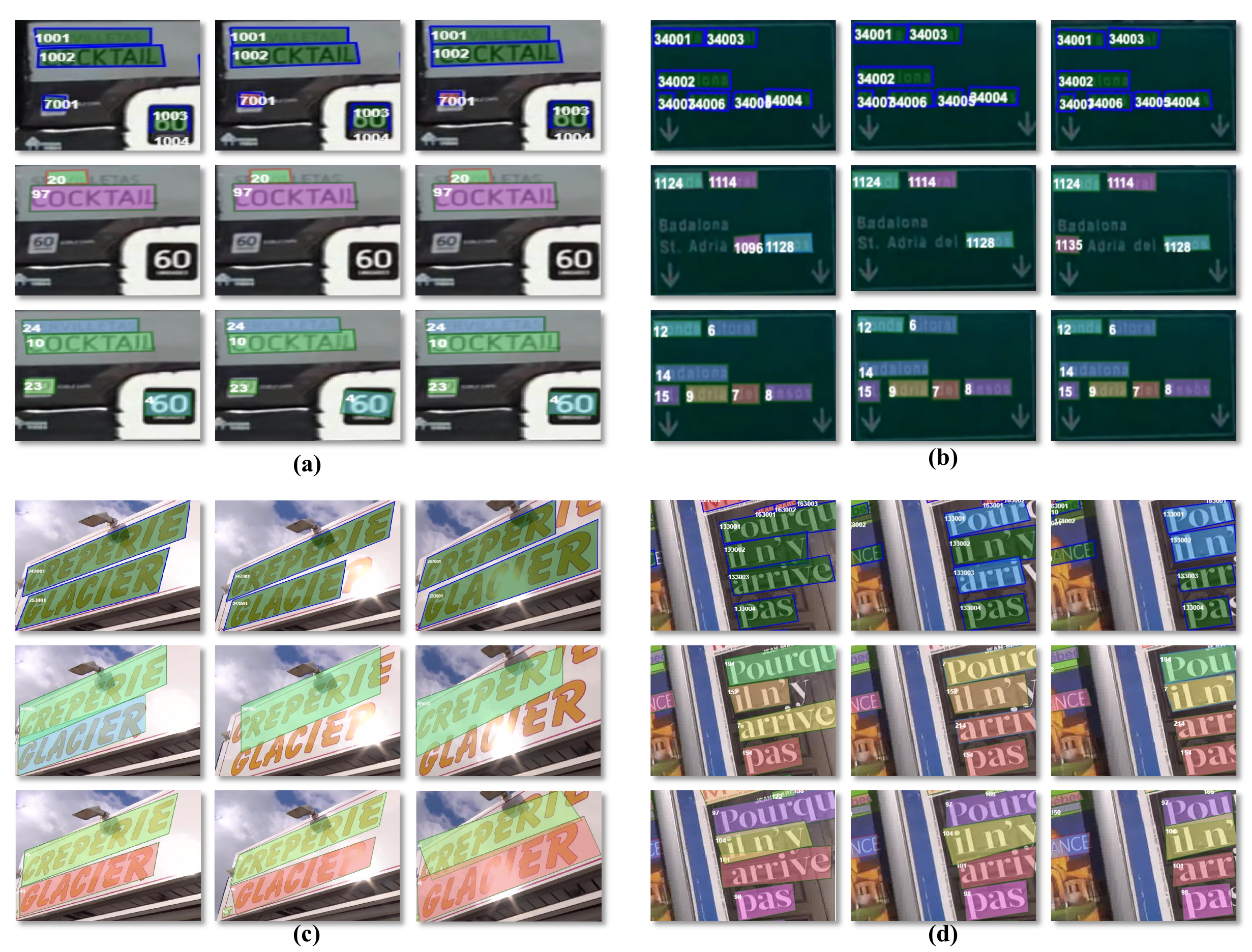}
    \caption{
    Qualitative results on the ICDAR 2015 Video Dataset. 
    (a), (b), (c), and (d) are from four different videos. For each video, images in the same row come from the consecutive video frames.
The top row is the ground-truth.
The middle row is the tracking results of HIK\_OCR \cite{cheng2021free(FREE)}.
The bottom row is the tracking results of ours. 
    The text instances with the same id are covered with masks of the same color. Figure best viewed in color.}
    \label{fig:qualitative_analysis}
\end{figure*}

\subsection{Method Analysis }
\subsubsection{Quantitative analysis }
We compare our method with the existing state-of-the-art approaches.
The experiment results on the ICDAR 2015 Video Text dataset and the Minetto dataset are shown in Table \ref{tab:comparison on tracking results icdar} and Table \ref{tab:comparison on tracking results minetto}, respectively. 
All the evaluation results in Table \ref{tab:comparison on tracking results icdar} are obtained from the ranking list on the ICDAR benchmark website.
Notably, in the ICDAR 2015 Video Text dataset, all the 1916 text trajectories in the videos of the testing set are classified into three categories (Mostly-Matched, Partial-Matched, Mostly-Lost) according to the evaluation results, as shown in Table \ref{tab:comparison on tracking results icdar}.

On the ICDAR 2015 video text dataset, our method achieves the best results on most evaluation metrics.
On the Minetto dataset, our method also outperforms other methods in all the evaluation metrics.
The superiority of our method mainly lies in two respects: 
On the one hand, 
The text instances embedding extracted by our text similarity learning network can help to efficiently distinguish the text instances from different trajectories. Hence we can correctly associate the text instances from the same trajectory, preventing the occurrence of ID switch.
As shown in Table \ref{tab:comparison on tracking results icdar}, our method obtains significant improvement in IDF1 compared with HIK\_OCR \cite{cheng2021free(FREE)}.
On the other hand, the proposed SCM is introduced to utilize the spatial-temporal relationship between the consecutive frames.
In this manner, the missed text can still be detected in some complex situations, such as motion blur and variation illumination. 
Hence, during text tracking, our SCM can effectively prevent the break of trajectories caused by misdetection.
As shown in Table \ref{tab:comparison on tracking results icdar}, the number of Mostly-Lost (ML) is decreased by 34.8$\%$ compared with HIK\_OCR \cite{cheng2021free(FREE)} (554 vs. 850).

We also present the evaluation results of text detection. 
Since there are no detection results reported on ICDAR 2015, detection results are all evaluated on ICDAR 2013 dataset.
The comparison results, presented in Table \ref{tab:detection result}, show that our method outperforms all other methods.
For instance, compared to the method of Yu et al. \cite{yu2019end(baidu)}, our method obtains improvements of 3.62$\%$ in F-measure (68.92 vs. 65.30) and 4.02$\%$ in recall (59.22 vs. 55.20), respectively.
The reason is that our SCM re-locates the text instances which are failed to be detected by the other methods.


\begin{table}[tp]
\renewcommand\arraystretch{1.4}
\caption{Text tracking results of our models with different feature embedding.}
\label{tab:ablation study for multi-modal embedding}
\begin{tabular}{ p{0.8cm}| p{1cm} | p{1cm} | p{0.8cm} | p{0.5cm} | p{0.4cm} | p{0.4cm} | p{0.5cm} }
\hline
Methods&$\text{MOTA}_D$$\uparrow$&$\text{MOTP}_D$$\uparrow$&IDF1$\uparrow$&MM$\uparrow$&PM$\downarrow$&ML$\downarrow$&ID Switch$\downarrow$ \\
\hline
VE &43.9$\%$&75.19$\%$ &\textbf{58.42}$\%$ & 858& 502& 556  &1126 \\
SE&42.42$\%$&75.19$\%$ &46.74$\%$ &858 &502 &556 &1861 \\
\hline
VE+SE&\textbf{44.07$\%$} &\textbf{75.19}$\%$ &58.23$\%$ &\textbf{858} &\textbf{502} &\textbf{556} &\textbf{1078}\\
\hline
\end{tabular}
\end{table}

\begin{table}[tp]
\renewcommand\arraystretch{1.2}
\centering
\caption{Text tracking results of our models with or without SCM.}
\label{tab:ablation study for sc}
\begin{tabular}{ p{1.2cm} | p{1cm} |  p{1cm} | p{0.8cm} | p{0.5cm} | p{0.5cm} | p{0.5cm} }
\hline
Methods&$\text{MOTA}_D$$\uparrow$&$\text{MOTP}_D$$\uparrow$&IDF1$\uparrow$&MM$\uparrow$&PM$\downarrow$&ML$\downarrow$ \\
\hline
w/o SCM &42.31$\%$ &\textbf{75.64}$\%$ &55.20$\%$ &734 &562 &620 \\
\hline
Ours & \textbf{44.07}$\%$&75.19$\%$ &\textbf{58.23}$\%$ &\textbf{858} &\textbf{502} &\textbf{556} \\
\hline
\end{tabular}
\end{table}

\begin{table}[tp]
\renewcommand\arraystretch{1.2}
\centering
\caption{Text detection results of our models with or without SCM.}
\label{tab:ablation study for sc detction}
\begin{tabular}{c|c|c|c}
\hline
Method&Precision&Recall&F-measure\\
\hline
w/o SCM &\textbf{84.55} &54.19 &66.05 \\
\hline
Ours &82.41&\textbf{59.22} &\textbf{68.92} \\
\hline
\end{tabular}
\end{table}

\subsubsection{Qualitative analysis}

Fig. \ref{fig:qualitative_analysis} shows the qualitative results on the ICDAR 2015 Video Text dataset of different scenes.
Specifically,  in the scenes of (a) and (b), the video frames are affected by motion blur. 
Obviously, In the middle row of scene (a), the text instance of trajectory ID 20 is partially-detected by HIK\_OCR \cite{cheng2021free(FREE)}.
In the middle row of scene (b), the text instance of trajectory ID 1096 is even missed.
However, our proposed SCM utilizes the spatial-temporal correlation between the consecutive frames to re-locate the missed text instances correctly and complete the partial-detected instances. 
Thus, the text trajectories keep complete in the results of our method.

In scene (c), the bottom text instance is truncated by intense illumination variation,
which results in missing detection of text instances by HIK\_OCR \cite{cheng2021free(FREE)}. Meanwhile, the complete text trajectories are obtained by our method.

\begin{figure}[t]
    \centering
    \includegraphics[width=0.85 \linewidth]{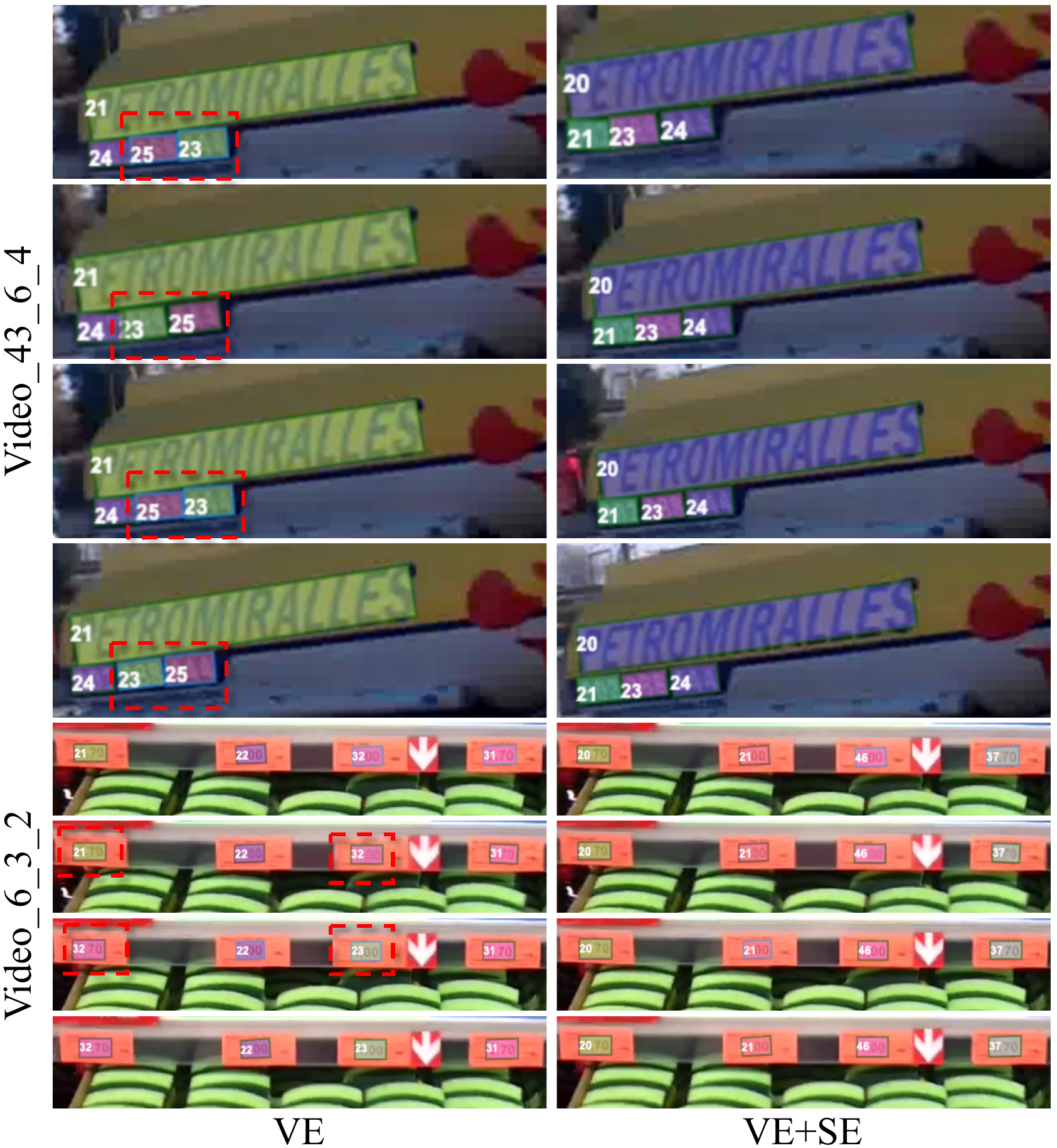}
    \caption{Visual comparison between VE and VE+SE. The red boxes indicate the ID switch.}
    \label{fig:VESE}
\end{figure}

{In scene (d), the text instances (denotes with trajectory IDs 133001,133002,133003) have very similar appearances, i.e., similar background and text font style.
As shown in the middle row of the scene (d),  
HIK\_OCR \cite{cheng2021free(FREE)} fails to distinguish these text instances,
and ID switch occurs in the tracking results.
By modeling the feature  embedding with both the visual and semantic information, 
our method can accurately distinguish these text instances. 
Thus, the correct trajectory IDs are assigned to each instance,
and ID switch is prevented.

\begin{table}[tp]
\renewcommand\arraystretch{1.2}
\centering
\caption{Text tracking results of our models with different association distances.}
\label{tab:ablation study for iou and morphology cost.}
\begin{tabular}{ p{1.3cm} | p{1cm} |  p{1cm} | p{0.8cm} | p{0.5cm} | p{0.5cm} | p{0.5cm} }
\hline
Methods&$\text{MOTA}_D$$\uparrow$&$\text{MOTP}_D$$\uparrow$&IDF1$\uparrow$&MM$\uparrow$&PM$\downarrow$&ML$\downarrow$ \\
\hline
only $\mathcal{D}_{e}$ &43.24$\%$ &75.19$\%$ &51.47$\%$ &858 & 502&556 \\
\hline
$\mathcal{D}_{e}$ + $\mathcal{D}_{m}$ &43.81$\%$ &75.19$\%$ &57.13$\%$ &858 &502 &556 \\
\hline
$\mathcal{D}_{e}$ + $\mathcal{D}_{p}$  &44.02$\%$ &75.19$\%$ &56.08$\%$ &858 &502 &556 \\
\hline
$\mathcal{D}_{e}$+$\mathcal{D}_{p}$+$ \mathcal{D}_{m}$  & \textbf{44.07}$\%$&\textbf{75.19}$\%$&\textbf{58.23}$\%$& \textbf{858}&\textbf{502} &\textbf{556} \\
\hline
\end{tabular}
\end{table}

\begin{figure}[tp]
    \centering
    \includegraphics[width=1 \linewidth]{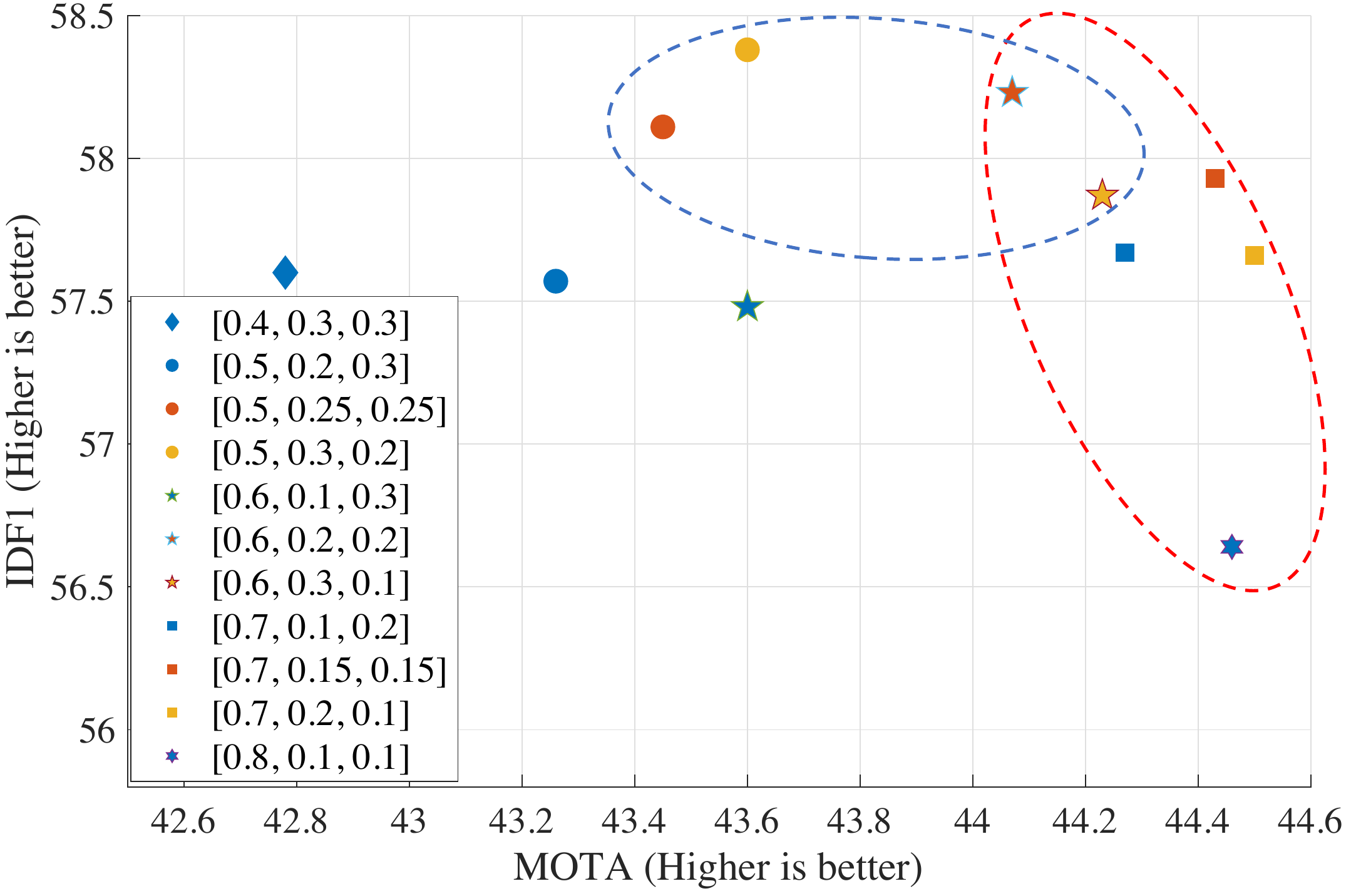}
    \caption{Comparison of model with different [$\alpha$, $\beta$, $\gamma$]. The red circle indicates the model achieving higher MOTA, blue for the model achieving higher IDF1.}
    \label{fig:abg}
\end{figure}

\subsection{Ablation Studies}
\subsubsection{Effectiveness of feature embedding}
A robust and discriminative feature embedding is important for text association. 
Our feature embedding is composed of visual embedding (VE) and semantic embedding (SE).
We perform experiments by removing the visual and semantic embedding.
The evaluation results are shown in Tab. \ref{tab:ablation study for multi-modal embedding},
The model with both two embeddings outperforms the model using VE or SE individually. 
In particular, using two embeddings obtains significant relative improvements of 4.3$\%$ (1078 vs. 1126) and 42.1$\%$ (1078 vs. 1861) in ID Switch metric over using SE and VE respectively.
The intrinsic reason is that our feature embedding is more discriminative by focusing on both the appearance and semantic information of the text instance.
Hence, we can distinguish the different text instances and correctly associate two instances from the same trajectory. 

Fig.\ref{fig:VESE} depicts the visual comparison between VE and VE+SE.
In video-43-6-4,
when only using visual embedding to represent text,
the text instances with ID 23 and 25 are always confused.
In video-6-3-2,
due to the similar appearance,
the text instances with ID 21, 32, and 23 are also confused.
In contrast,
the model using VE and SE together addresses this issue better.

\subsubsection{Effectiveness of SCM} 
In order to verify the effectiveness of our proposed SCM,
we perform the comparative experiments with or without the SCM.
The tracking results and detection results are shown in Tab. \ref{tab:ablation study for sc} and Tab. \ref{tab:ablation study for sc detction}, respectively. 
For text tracking, using the SCM achieves the improvement of $1.76\%$ in $\text{MOTA}_D$ and $3.03\%$ in IDF1 over without the SCM.
For text detection, using the SCM outperforms by $2.87\%$  in F-measure compared to without the SCM.
In particular, the SCM makes a great contribution to recall by $5.03\%$.  
Intuitively, our SCM re-locates the missing text instances by leveraging the spatial-temporal relevance between the consecutive frames, thus boosting the recall of text detection.

\subsubsection{Effectiveness of similarity measure}}

In Tab. \ref{tab:ablation study for iou and morphology cost.}, 
we compare performance among the embedding distance $\mathcal{D}_{e}$, the IOU distance $\mathcal{D}_{p}$, and morphology distance $\mathcal{D}_{m}$.
The evaluation results reveal that the performances are boosted after utilizing either the $\mathcal{D}_{p}$ or $\mathcal{D}_{m}$.
Notably, the IDF1 is significantly improved, which proves that these two distance metrics can reduce the ID switch effectively. 
Thus, these evaluation results confirm the effectiveness of our design. 

Intuitively, two text instances that belong to the same text trajectory have a similar location. The bounding boxes of these text instances also have more similar morphology shape. 
Thus, the IOU distance $\mathcal{D}_{p}$ and the morphology distance $\mathcal{D}_{m}$ help to distinguish different text instances while making text association more robust.

\subsection{Analysis for Parameters}
\label{sec:param}
Table \ref{tab:h1h2.} depicts the performance of tracking with different $h_1$ and $h_2$.
It can be seen that the performance fluctuated slightly when $ h_1 $ and $ h_2 $ changed.
In more detail, $ h_2 $ affects the performance of the model to a greater extent than $ h_1 $.
Furthermore, the model with $h_1$ of 0.6 and $h_2$ of 0.3 obtains the best performance on MOTA and MOTP and the competitive performance on IDF1, MM, PM and ML.
Thus, we argue that (0.6, 0.3) is the optimal parameter setting for $h_1$ and $h_2$.

For the weights $\alpha$, $\beta$, $\gamma$ of different distances,
since the appearance distance $\mathcal{D}_e$ is dominant in three types of distance,
we first set $\alpha$ to different values,
then change the value of $\beta$ and $\gamma$.
Fig. \ref{fig:abg} demonstrates their performance.
When $\alpha$ is set to 0.5 and 0.6,
the model achieves a better performance on IDF1.
When $\alpha$ is set to 0.6, 0.7, 0.8,
the model achieves a better performance on MOTA.
In addition,
it would be better to set the same value to $\beta$ and $\gamma$.
Therefore,
we set the three weights as $\alpha=0.6$, $\beta=0.2$, $\gamma=0.2$.

\begin{figure}
    \centering
    \includegraphics[width=1 \linewidth]{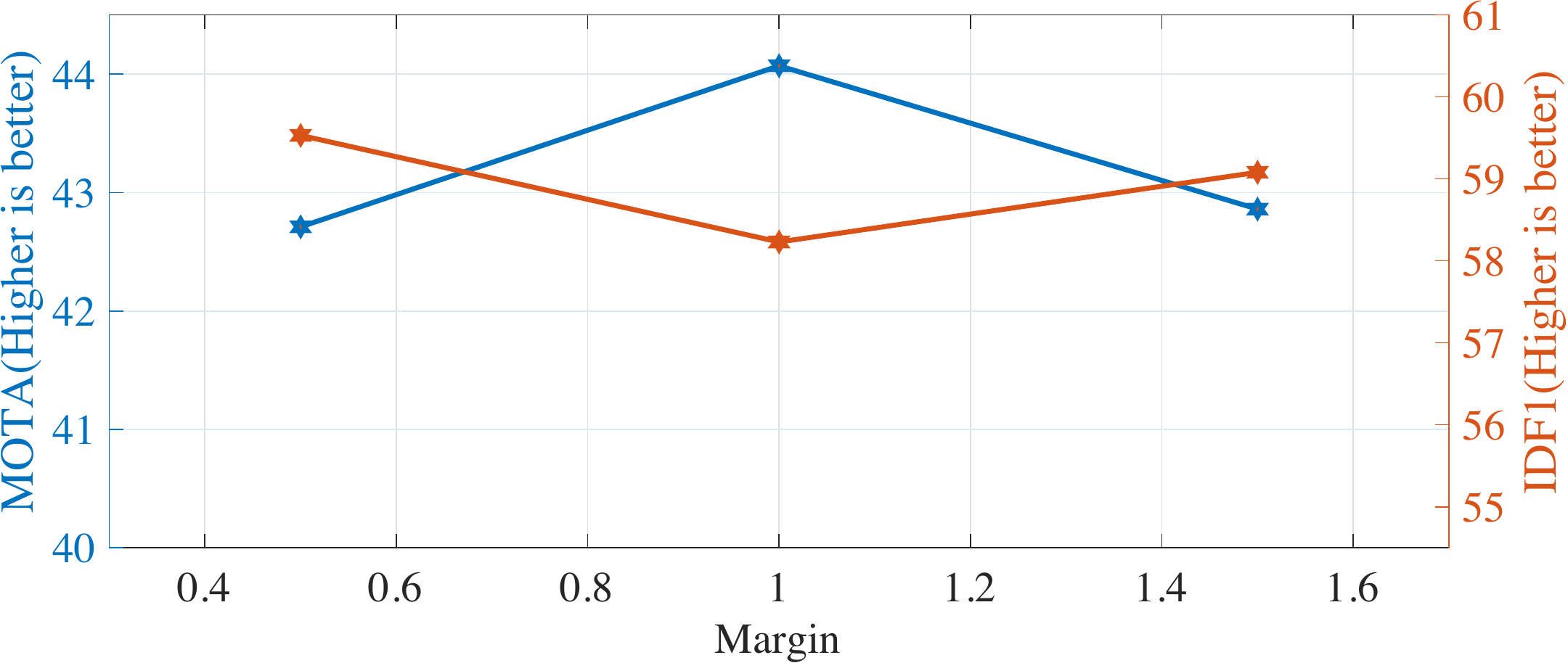}
    \caption{Comparison of model with different margins in triplet loss.}
    \label{fig:M}
\end{figure}

For parameter $\mathcal{M}$,
Fig. \ref{fig:M} depicts the results with different margins $\mathcal{M}$ of triplet loss.
It can be seen that the model with $\mathcal{M}$ of 1 obtains the highest MOTA and a low IDF1.
Since the improvement on MOTA is higher than the decrease on IDF1 when $\mathcal{M}=1$,
we set $\mathcal{M}=1$ in our network.

\begin{table}[tp]
\renewcommand\arraystretch{1.2}
\centering
\caption{Text Detection Results with different setting of detection parameters $h_1$ and $h_2$.}
\label{tab:h1h2.}
\begin{tabular}{p{0.4cm} | p{0.4cm} | p{1cm} |  p{1cm} | p{0.8cm} | p{0.5cm} | p{0.5cm} | p{0.5cm}}
\hline
$h_1$&$h_2$&$\text{MOTA}_D$$\uparrow$&$\text{MOTP}_D$$\uparrow$&IDF1$\uparrow$&MM$\uparrow$&PM$\downarrow$&ML$\downarrow$ \\
\hline
0.6& 0.3& \textbf{44.07}$\%$& \textbf{75.19}$\%$& 58.23$\%$& 858& 502& 556 \\
\hline
0.6& 0.4& 43.15$\%$& 74.88$\%$& 58.57$\%$& 916& 478& 522 \\
\hline
0.5& 0.3& 44.00$\%$& 75.15$\%$& 57.99$\%$& 873& 485& 558 \\
\hline
0.5& 0.4& 42.73$\%$& 74.85$\%$& \textbf{58.61}$\%$& \textbf{920}& \textbf{477}& \textbf{519} \\
\hline
\end{tabular}
\end{table}

\begin{table}[tp]
\renewcommand\arraystretch{1.2}
\centering
\caption{Inference time of different modules.}
\label{tab:speed.}
\begin{tabular}{c | c | c }
\hline
Steps & Modules in our model & Inference Time /ms \\
\hline
\multirow{3}{*}{Detection}& Text detector& 12.864 \\
\cline{2-3}
& SCM & 1110.772 \\
\cline{2-3}
& Mask fusion & 22.193 \\
\hline
\multirow{3}{*}{Association}& Similarity learning network & 5.560\\
\cline{2-3}
& Associating text & 3.862 \\
\cline{2-3}
& Trajectory updating & 1.762 \\
\hline
\end{tabular}
\end{table}


\subsection{Speed Analysis}
Speed is also an essential factor in evaluating text tracking.
With the computation platform of NVIDIA TITAN Xp,
the inference time per module can be seen in Table \ref{tab:speed.}.
For our method,
the computation mainly consists of two parts: detection and association.
In the detection step,
the text detector takes 12.864 ms,
the mask fusion takes 22.193 ms,
and our siamese complementary module (SCM) spends 1110.772 ms,
which takes up most of the time.
The reason is that the number of feature extraction is equal to the number of text instances $d^{t-1}_i$,
leading to the redundant feature extraction in search region $\mathcal{X}^t_i$.
In this case, the time complexity can be effectively reduced by integral map.
In the future,
we are going to optimize the time-consuming of SCM by simplifying the feature extraction.
In addition, the association step takes 11.184 ms in total. 


\yz{
\section{Limitation and Future Work}

As above-mentioned, the main limitation of our approach is the inference time of Siamese Complementary Module (SCM), 
since it contains redundant feature extraction \xingl{process.}
In the future, we will try to improve the parallelization of SCM to speed up the whole system.
In addition, we will try to integrate the detection module and the matching module,
i.e., SCM, into a unified end-to-end trainable network, thereby reducing unnecessary hyper-parameters and enhancing the robustness of the model.
}

\section{Conclusion}
This paper presents a novel video text tracking model to resolve the main challenges in video text tracking. 
We propose the SCM to find the missed text instance by leveraging the relevance between consecutive frames.
Besides, a text similarity learning network is designed to learn and output the feature embedding of each text instance. Further, the visual embedding and semantic embedding are combined to form a more discriminative embedding for text association. 
The quantitative and qualitative analysis on publicly available datasets demonstrates the effectiveness of our method and the designed modules. Our method also obtains state-of-the-art performance on public benchmarks, which further verifies the superiority of our method.


\ifCLASSOPTIONcaptionsoff
  \newpage
\fi


\bibliographystyle{IEEEtran}
\bibliography{IEEEabrv,IEEEexample}
\begin{IEEEbiography}[{\includegraphics[width=1in,height=1.25in,clip,keepaspectratio]{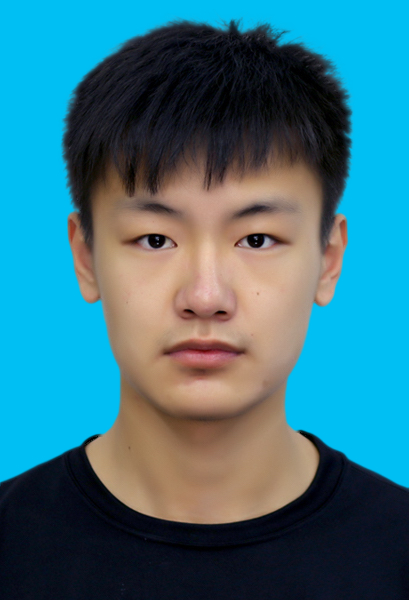}}]{Yuzhe Gao} received the M.S. degree from Huazhong University of Science and Technology, Wuhan, China. His reaserch interests include computer vision, scene understanding and occlusion relationship reasoning.
\end{IEEEbiography}

\begin{IEEEbiography}[{\includegraphics[width=1in,height=1.25in,clip,keepaspectratio]{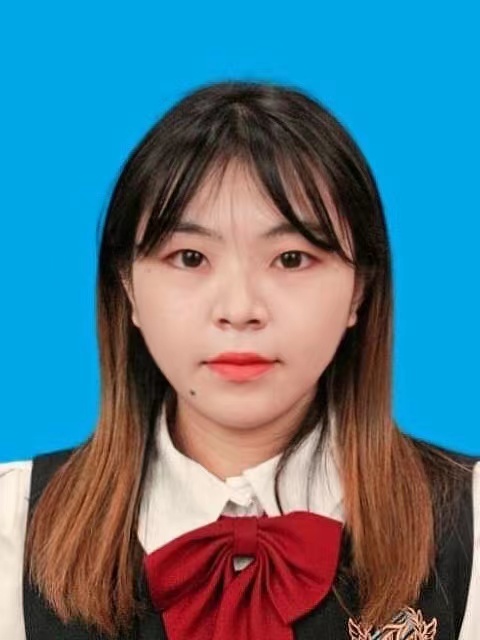}}]{Xing Li} received her B.S. degree in Huazhong University of Science and Technology (HUST).She is currently pursuing the M.S. degree from School of Electronic Information and Communications, HUST. Her research interests mainly focus on computer vision, including text recognition, detection and text tracking. 
\end{IEEEbiography}
\begin{IEEEbiography}[{\includegraphics[width=1in,height=1.25in,clip,keepaspectratio]{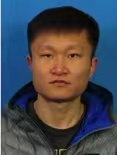}}]{Jiajian Zhang} graduated from Huazhong University of Science and Technology, Wuhan, China. His research topic is computer vision. He mainly researched in video Text Tracking when he was a master student in School.
\end{IEEEbiography}
\begin{IEEEbiography}[{\includegraphics[width=1in,height=1.25in,clip,keepaspectratio]{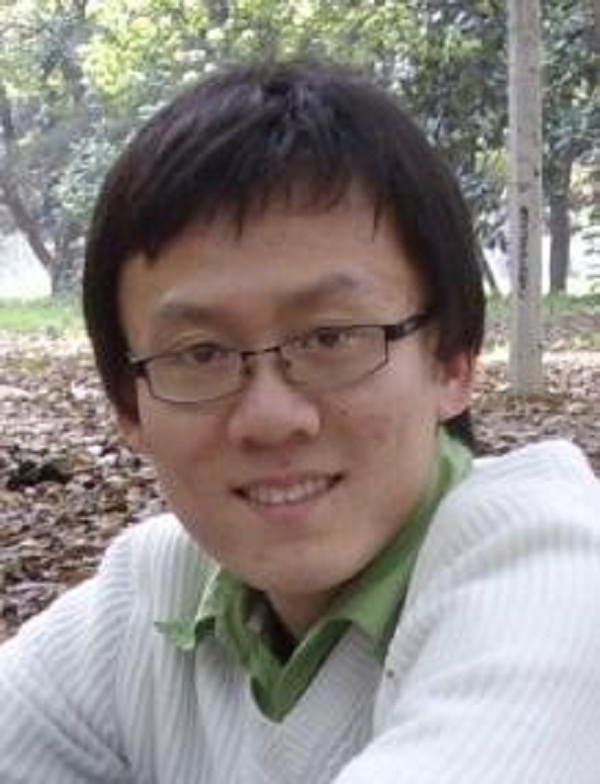}}]{Yu Zhou} received the M.S. and Ph.D. degrees both in Electronics and Information Engineering from Huazhong University of Science and Technology (HUST), Wuhan, P.R. China in 2010, and 2014, respectively.
In 2014, he joined the Beijing University of Posts and Telecommunications (BUPT), Beijing, as a Postdoctoral Researcher from 2014 to 2016, an Assistant Professor from 2016 to 2018.
He is currently an Associate Professor with the School of Electronic Information and Communications, HUST.
His research interests include computer vision and automatic drive.
\end{IEEEbiography}
\begin{IEEEbiography}[{\includegraphics[width=1in,height=1.25in,clip,keepaspectratio]{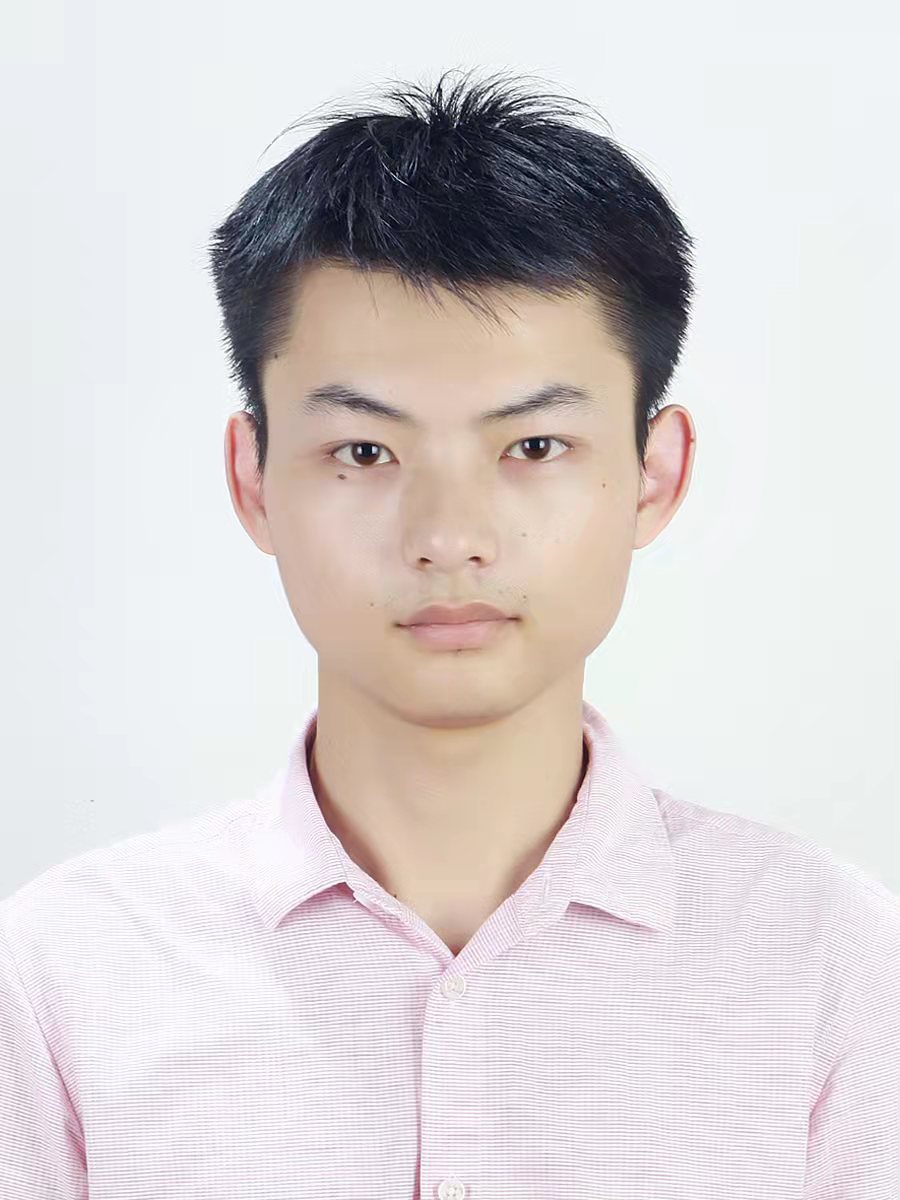}}]{Dian Jin} received his B.S degree and M.S degree from the school of Electronic information and Communications, 
Huazhong University of Science and Technology(HUST), China respectly in 2018 and 2021. His main research interests include scene text detection and recognition
\end{IEEEbiography}
\begin{IEEEbiography}[{\includegraphics[width=1in,height=1.25in,clip,keepaspectratio]{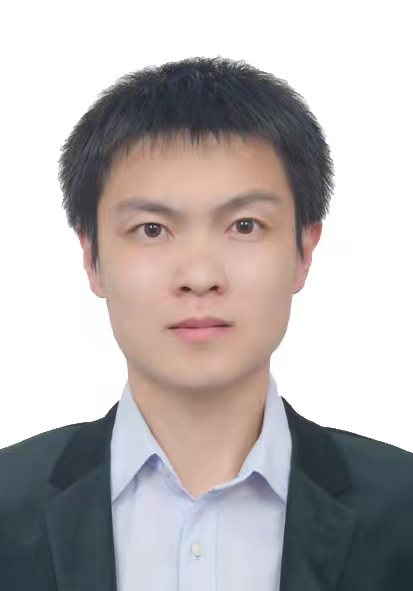}}]{Wang Jing} is a senior AI algorithm engineer for Huawei Cloud OCR. He has many years of algorithm experience and has obtained his doctorate and bachelor's degree in mathematics and applied mathematics from Nanyang Technological University in Singapore and University of Science and Technology of China, respectively. He is responsible for the core algorithm of text recognition, and has submitted multiple deep learning-based text detection and recognition patents and papers. He is familiar with cloud computing, artificial intelligence, cryptography and computer network security.
\end{IEEEbiography}
\begin{IEEEbiography}[{\includegraphics[width=1in,height=1.25in,clip,keepaspectratio]{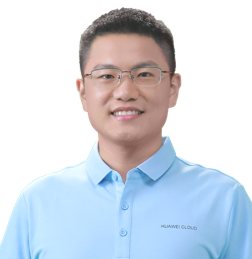}}]{Shenggao Zhu}received the Ph.D. degree in Computer Science from National University of Singapore (NUS), 2017. He got his bachelor in Electronic Engineering and Information Science from University of Science and Technology of China (USTC), 2011. He joined Huawei Cloud in 2017 and now is a Technical Expert. His research interests include computer vision and AI applications.
\end{IEEEbiography}

\begin{IEEEbiography}[{\includegraphics[width=1in,height=1.25in,clip,keepaspectratio]{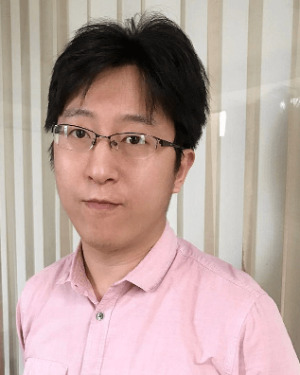}}]{Xiang Bai} received his B.S., M.S., and Ph.D. degrees from the Huazhong University of Science and Technology (HUST), Wuhan, China, in 2003, 2005, and 2009, respectively, all in electronics and information engineering. He is currently a Professor with the School of Artiﬁcial Intelligence and Automation, HUST. His research interests include object recognition, shape analysis, and OCR. He has published more than 150 research papers. He is an editorial member of IEEE TPAMI, Pattern Recognition, and Frontier of Computer Science. He is the recipient of 2019 IAPR/ICDAR Young Investigator Award for his outstanding contributions to scene text understanding. He is a senior member of IEEE.
\end{IEEEbiography}
\end{document}